\documentclass[sigconf]{acmart}
\usepackage{multirow}
\usepackage{enumitem}
\usepackage{subcaption}
\usepackage{algorithm}
\usepackage{algorithmic}

\copyrightyear{2025}
\acmYear{2025}
\setcopyright{acmlicensed}
\acmConference[MM '25]{Proceedings of the 33rd ACM International Conference on Multimedia}{October 27--31, 2025}{Dublin, Ireland}
\acmBooktitle{Proceedings of the 33rd ACM International Conference on Multimedia (MM '25), October 27--31, 2025, Dublin, Ireland}
\acmDOI{10.1145/3746027.3755809}
\acmISBN{979-8-4007-2035-2/2025/10}

\acmSubmissionID{6136}

\begin{document}

\ccsdesc[500]{Information systems~Multimedia information systems}
\settopmatter{printacmref=true}

\title{Consistent and Invariant Generalization Learning for Short-video Misinformation Detection}



\author{Hanghui Guo}
\authornotemark[1]
\authornotemark[3]
\affiliation{%
  \institution{Zhejiang Normal University}
  \state{Zhejiang}
  \country{China}
}
\email{ghh1125@zjnu.edu.cn}

\author{Weijie Shi}
\authornote{These authors contributed equally to this work.}
\affiliation{%
  \institution{Hong Kong University of Science and Technology}
  \state{Hong Kong SAR}
  \country{China}}
\email{wshiah@connect.ust.hk}

\author{Mengze Li}
\authornotemark[2]
\affiliation{%
  \institution{Hong Kong University of Science and Technology}
  \state{Hong Kong SAR}
  \country{China}}
\email{mengzeli@zju.edu.cn}

\author{Juncheng Li}
\affiliation{%
 \institution{Zhejiang University}
  \state{Zhejiang}
 \country{China}}

\author{Hao Chen}
\affiliation{%
  \institution{Tencent}
  \country{China}}

\author{Yue Cui}
\affiliation{%
  \institution{Hong Kong University of Science and Technology}
  \state{Hong Kong SAR}
  \country{China}}

\author{Jiajie Xu}
\affiliation{%
  \institution{Soochow University}
    \state{Jiangsu}
  \country{China}}

\author{Jia Zhu}
\authornote{Corresponding author.}
\authornote{Hanghui Guo and Jia Zhu are with the Zhejiang Key Laboratory of Intelligent Education Technology and Application, Zhejiang Normal University.}
\affiliation{%
  \institution{Zhejiang Normal University}
    \state{Zhejiang}
  \country{China}}
\email{jiazhu@zjnu.edu.cn}

\author{Jiawei Shen}
\affiliation{%
  \institution{Zhejiang Normal University}
    \state{Zhejiang}
  \country{China}}

\author{Zhangze Chen}
\affiliation{%
  \institution{Zhejiang Normal University}
  \state{Zhejiang}
\country{China}}
    
\author{Sirui Han}
\affiliation{%
  \institution{Hong Kong University of Science and Technology}
  \state{Hong Kong SAR}
  \country{China}}
\renewcommand{\shortauthors}{Hanghui Guo et al.}



\begin{abstract}
Short-video misinformation detection has attracted wide attention in the multi-modal domain, aiming to accurately identify the misinformation in the video format accompanied by the corresponding audio. Despite significant advancements, current models in this field, trained on particular domains (source domains), often exhibit unsatisfactory performance on unseen domains (target domains) due to domain gaps. To effectively realize such \textit{domain generalization} on the short-video misinformation detection task, we propose deep insights into the characteristics of different domains: (1) The detection on various domains may mainly rely on different modalities (\emph{i.e.}, mainly focusing on videos or audios). To enhance domain generalization, it is crucial to achieve optimal model performance on all modalities simultaneously. (2) For some domains focusing on cross-modal joint fraud, a comprehensive analysis relying on cross-modal fusion is necessary. However, domain biases located in each modality (especially in each frame of videos) will be accumulated in this fusion process, which may seriously damage the final identification of misinformation. To address these issues, we propose a new DOmain generalization model via ConsisTency and invariance learning for shORt-video misinformation detection (named \textbf{DOCTOR}), which contains two characteristic modules: (1) We involve the cross-modal feature interpolation to map multiple modalities into a shared space and the interpolation distillation to synchronize multi-modal learning; (2) We design the diffusion model to add noise to retain core features of multi modal and enhance domain invariant features through cross-modal guided denoising. Extensive experiments demonstrate the effectiveness of our proposed DOCTOR model. Our code is publicly available at \url{https://github.com/ghh1125/DOCTOR}.
\end{abstract}



\keywords{Misinformation Detection; Domain Generalization; Multi-modal}



\maketitle

\section{Introduction}
With the rapid rise of short video platforms like TikTok, misinformation in the short-video format accompanied by matching audio is spreading rapidly and widely \cite{zhang2023hierarchical, zong2024unveiling, bu2024fakingrecipe, agarwal2024television}. Such short-video misinformation may seriously mislead public decision-making and behavior, which poses a potential threat to social stability \cite{xu2023combating, ali2022effects}.

\begin{figure}
    \centering
    \includegraphics[width=0.9\linewidth]{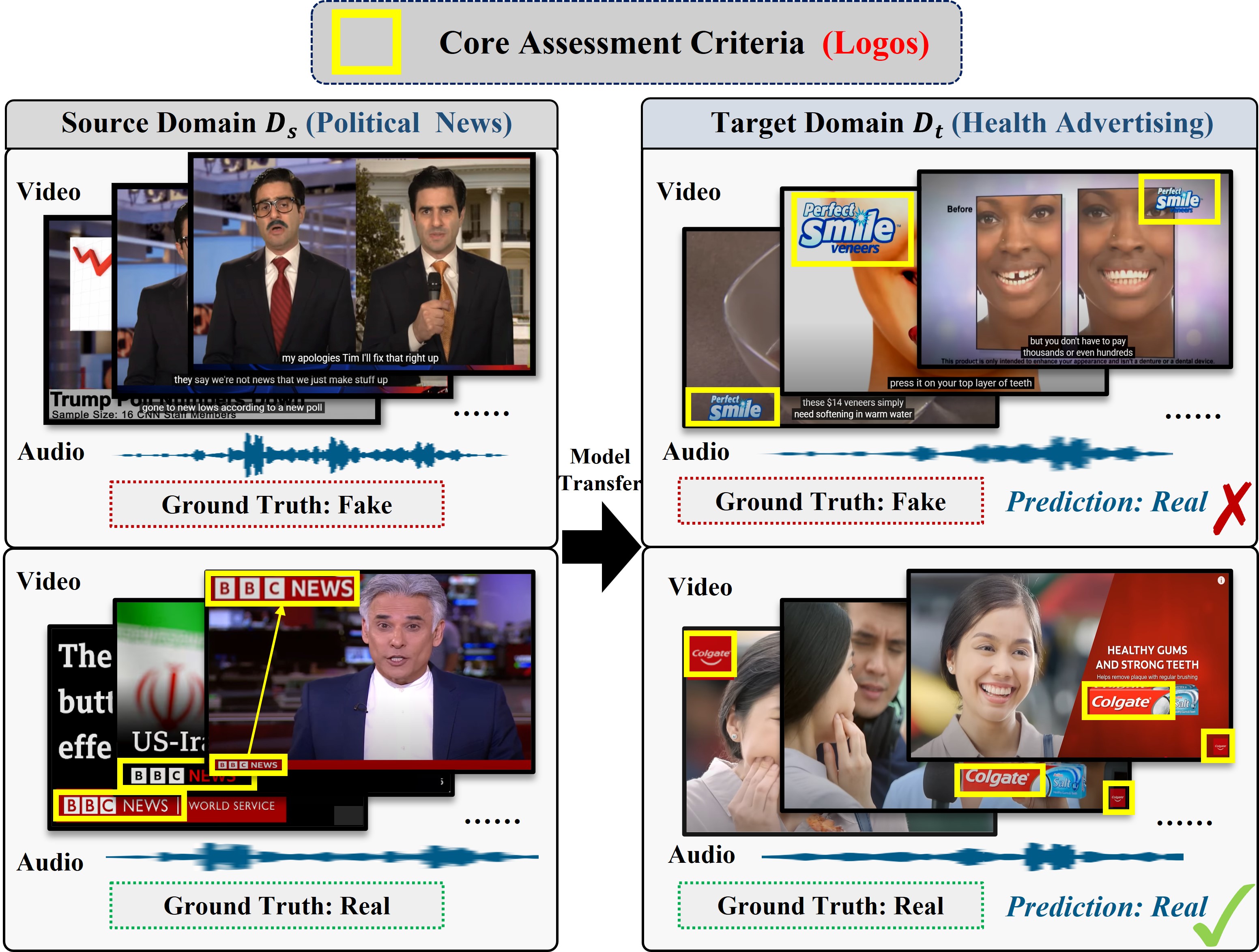}
    \caption{An example of how \textit{\color{red}domain bias (like logos)} can lead to errors in misinformation detecting of the target domain.}
    \label{fig:Domain}
\end{figure}

Hence, there is an urgent requirement to develop effective methods for promptly identifying misinformation to control its spread \cite{pang2022tackling}. 
Benefiting from the development of deep learning technology, existing methods achieve promising results in this field \cite{zhang2024early, bu2024fakingrecipe}. 
Nevertheless, when trained on specific domains (source domains $\mathcal{D}_s$), these data-driven methodologies experience a notable decline in performance when directly being applied to unseen domains (target domains $\mathcal{D}_t$) \cite{wang2024search}. The failure of such \textit{domain generalization} is primarily attributed to the existence of substantial differences in data distribution across various domains.
Figure \ref{fig:Domain} illustrates an example where, in the political news domains, many logos mean information coming from official media, which may represent its authenticity; whereas, in health advertising domains, almost all short videos (\emph{i.e.}, real and fake ones) are accompanied by product logos. This shift of \textit{domain bias} leads to detection errors of misinformation.

\begin{figure}
    \centering
    \includegraphics[width=0.9\linewidth]{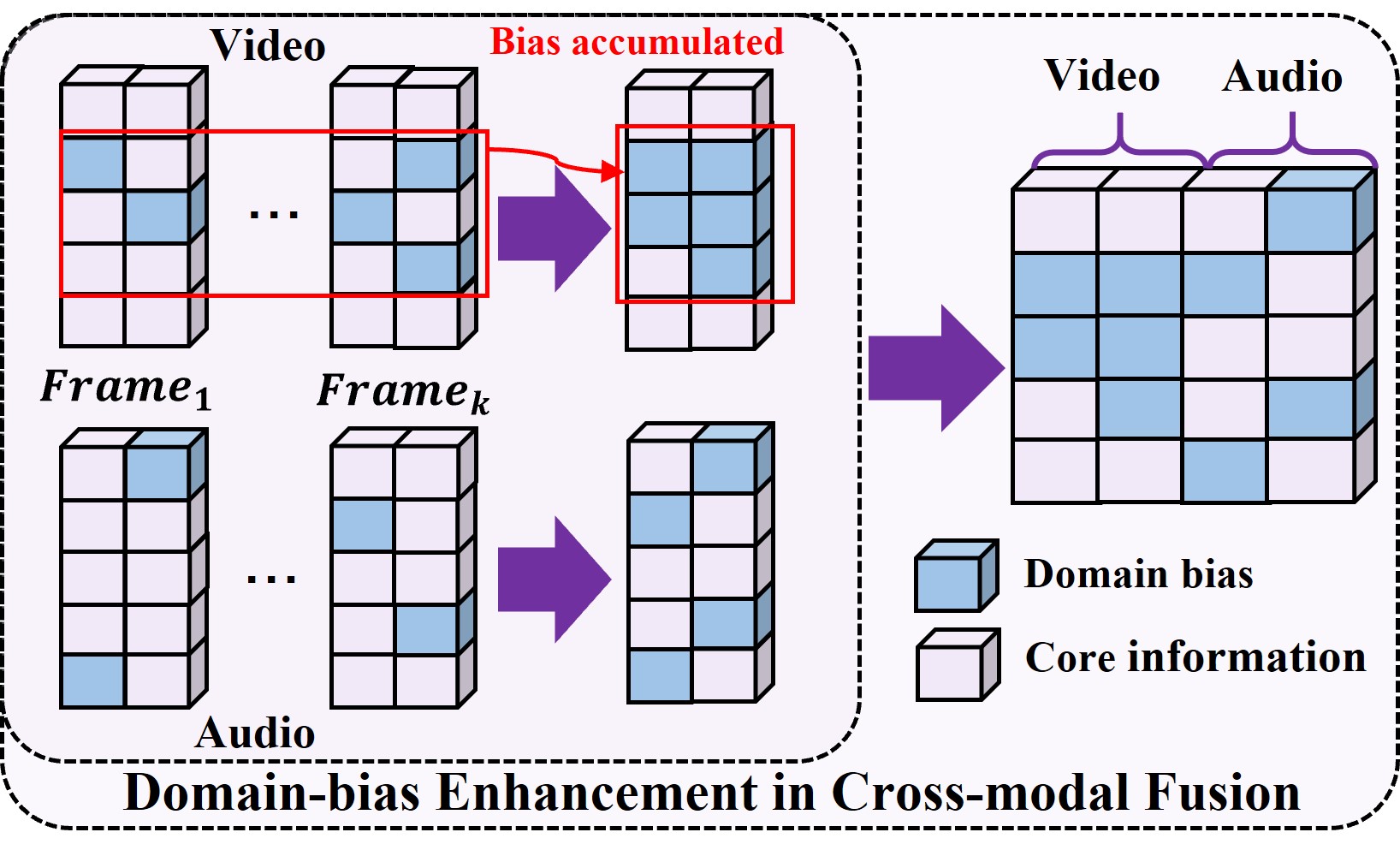}
    \caption{The problem of domain-bias Enhancement in Cross-modal Fusion. The red line is a case of \textit{bias accumulated}.}
    \label{fig:1}
\end{figure}

To effectively realize the domain generalization on the short-video misinformation detection task, we provide deep insights into the characteristics of different domains. 
In particular, various domains exhibit distinct forms of fraud, encompassing activities such as news falsification focused on a single modality as well as the convergence of multiple modalities for deceptive purposes. 
Due to the existing domain gap, designing methods to generalize to diverse domains will encounter different problems: 

\textbf{(1) Varying Modal-dependence in Different Domains.} In cases where particular modal information in short-video news is falsified (like only falsifying video or audio), various domains may prioritize different types of modalities for their focus \cite{jin2022evaluating, bu2024fakingrecipe, liu2024continual}.
To fully enhance the domain generalization, the reasoning ability in any modality should be fully optimized when training misinformation detection models \cite{qi2021improving, zhang2024reinforced}. However, during such cross-modal joint learning, multiple modalities may compete with each other. This may ultimately lead to insufficient knowledge acquisition for parts of modalities,
making it challenging to achieve the sweet point on all modalities simultaneously.

\textbf{(2) Domain-bias Enhancement in Cross-modal Fusion.} 
For news falsification on multiple modalities together, a comprehensive analysis based on effective cross-modal fusion is necessary for accurate falsification detection \cite{li2021entity, hu2024cross, yang2023deep}. 
However, as detailed shown in Figure~\ref{fig:1}, simply summing the features of all modalities will add up the domain bias located in the features of each modality.
In particular, video features introduced may be accompanied by a significant accumulation of domain bias stemming from the numerous non-core video frames and the non-core frame areas involved.
Such bias accumulation from all modalities seriously highlights bias impact in the obtained feature and damages the final detection.

To tackle the aforementioned issues, we introduce a new DOmain generalization model via ConsisTency and invariance learning for shORt-video misinformation detection (named \textbf{DOCTOR}). Specifically, there are two specific designs: 
\textbf{(1) Cross-modal Interpolation Distillation.} 
We project video and audio modality features into a shared representation space, and mix these representations with a random mixing ratio to construct interpolations. Based on the feature distribution of these interpolation samples as teacher signals, feature distillation drives the learning optimization process of the student model, making the representation learning across modalities more consistent and synchronized.
\textbf{(2) Multi-modal Invariance Fusion.} 
Incorporating subtitles extracted from each video frame as a guide, we devise a novel mechanism to facilitate the extraction of core video features from both spatial and temporal viewpoints.
With extracted multi-modal features, we add noise to them step by step based on the diffusion model to filter out redundant details, thereby underlining the core features of multi-modal. Subsequently, by gradually mitigating the noise influence, we further outline the fundamental information of multi-modal features through cross-modal mutual guidance, boosting domain invariant features. 
The main contributions of our work are as follows:

\begin{itemize}[leftmargin=*]   
    \item To the best of our knowledge, we take the early exploration of the domain generalization on the short-video misinformation detection task. For this new setting, we introduce the DOCTOR model via consistency and invariance learning.
    \item We propose distillation for the cross-modal feature interpolation to coordinate model learning of multiple modalities. In addition, we involve diffusion-based domain bias filtering and invariant feature learning by step-by-step adding noise and denoising.
    \item We validate the effectiveness of our DOCTOR model through extensive comparative experiments.
\end{itemize}

\begin{figure*}
    \centering
    \includegraphics[width=\textwidth]{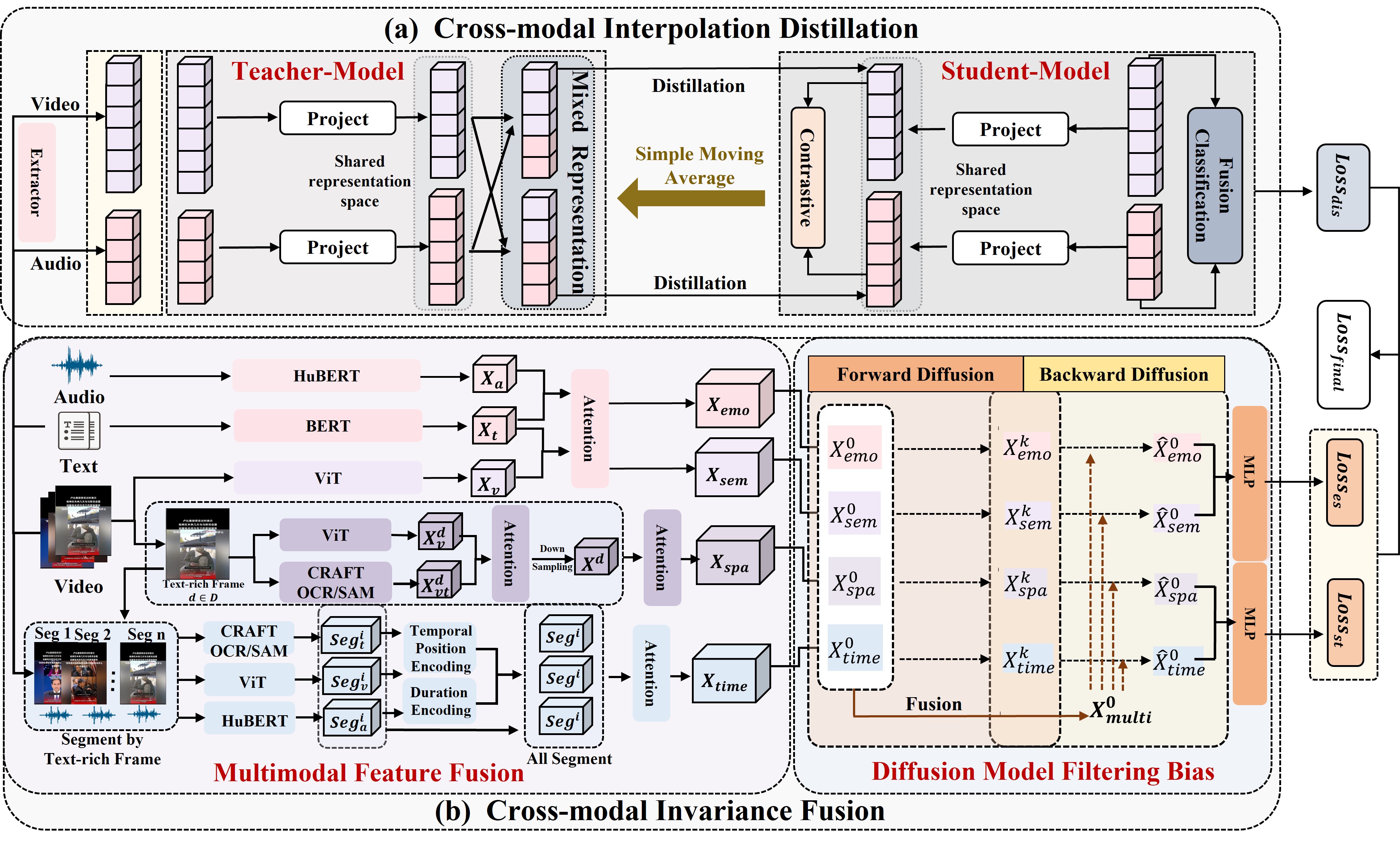}
    \caption{The framework of the DOCTOR. (a) Cross-modal Interpolation Distillation: Video and audio features are projected into a shared representation space and mixed with a random ratio to create interpolation samples. These samples are then used as teacher signals to guide the student model in learning consistent representations across modalities. (b) Cross-modal Invariance Fusion: Multi-modal features are first fused, then progressively perturbed through a diffusion model to mask domain biases. The noise is gradually removed to enhance core multi-modal features, aiming to obtain domain-invariant features.}
    \label{fig:DOCTOR}
\end{figure*}

\section{Related Works}

\subsection{Multi-modal Misinformation Detection}

Early research on multi-modal misinformation detection primarily focuses on text and image, emphasizing cross-modal correlation \cite{zhao2023enhancing, zhang2023hierarchical, liu2024fka, zhang2024mitigating, wang2023cross, ma2024cross}. For example, MCAN \cite{wu2021multimodal} designs co-attention networks to fuse textual and visual features. CAFE \cite{chen2022cross} transforms features into a shared semantic space through cross-modal semantic alignment and combines cross-modal to estimate the fuzziness between different modalities. MMICF \cite{zeng2023correcting} eliminates the direct disparity in the fusion between text and image entities through causalities. Although these methods have shown promising performance in detecting misinformation involving text and images, their direct applicability to short-video misinformation is limited due to fundamental differences in content mechanisms \cite{bu2023combating}.

With the rise of online short-video platforms, the spread of misinformation has expanded to video-audio. Some short-video misinformation detection methods have been proposed through recent research. OpEvFake \cite{zong2024unveiling} employs a large language model to mine implicit viewpoints in short videos and combines model information to enhance interaction for judgment. SV-FEND \cite{qi2023fakesv} exploits cross-modal correlations to select the most informative features and uses social context information for detection. FakingRecipe \cite{bu2024fakingrecipe} detects misinformation by simulating the creation process of audio and video. TikTec \cite{shang2021multimodal} employs speech text-guided visual object features and MFCC-guided speech textual features. However, these methods are typically designed for general domains and often underperform in domain generalization scenarios.

\subsection{Multi-modal Domain Generalization}

While uni-modal domain generalization has been well-studied (such as only text, image, or video), research on video-audio domain generalization remains limited \cite{ouali2023black, xu2022ava, oh2023geodesic}. Some of the recent progress on video-audio includes: SimMMDG \cite{dong2023simmmdg} increases the distance between modality-specific features and modality-shared features of audio and video. CMRF \cite{fan2024cross} improves domain generalization by reducing the flat minima between different modalities. RNA-Net \cite{planamente2022domain} balances audio and video feature norms via a relative norm alignment loss. However, the existing domain generalization focus on video-audio methods is primarily tailored for general domains rather than specifically designed for misinformation detection, resulting in limited task adaptability and suboptimal performance.

In contrast, domain generalization for misinformation has currently focused on text-image modalities, using domain-aware fusion and representation learning \cite{castelo2019topic, silva2021embracing, zhu2022memory, tong2024mmdfnd}. For instance, EDDFN \cite{silva2021embracing} learns features from domain-specific and cross-domain embeddings and then connects domain-related knowledge. MDFEND \cite{nan2021mdfend} aggregates multiple representations extracted by experts through domain gates. MMDFND \cite{tong2024mmdfnd} integrates cross-domain and domain-specific knowledge through the design of expert networks and step-by-step pivot transformers. However, these methods rely on static alignment, which is inadequate for video-audio data that involve temporal dynamics and complex interactions \cite{zhu2024vision+, wang2025cross}. Therefore, bridging the domain generalization gap in short-video misinformation detection remains an urgent and underexplored challenge.

\section{Proposed Method: DOCTOR}

In this section, we propose an innovative DOmain generalization model via ConsisTency and invariance learning for shORt-video misinformation detection (named \textbf{DOCTOR}).

\subsection{Model Overview}

\subsubsection{Problem Formulation} \textbf{Short-video misinformation detection.}
We assume that the input is a short video $v$ and accompanying audio $a$ (sometimes matched with a text description $t$ about the video title or introduction). After judging whether it contains misinformation, the output predicted label is represented as $Y \in \{0,1\}$, where $0$ represents misinformation and $1$ represents the opposite. Building a misinformation detection model $\mathcal{M}$, input multi-modal data $\{v,a,t\}$, output predicted labels $\hat{Y}$, we minimize the loss function $\mathcal{L}(\cdot)$ between the predicted label $\hat{Y}$ and the true label $Y$ by optimizing the parameter $\Theta$. The following is the target formula:
\begin{equation}
    \min_{\Theta} \mathbb{E}_{(v, a, t, Y) \sim D} \left[ \mathcal{L}(Y, \hat{Y}) \right], \quad \hat{Y} = \mathcal{M}(v, a, t; \Theta),
\end{equation}
where $D$ is the joint distribution over all inputs and labels.

\textbf{Domain generalization for short-video misinformation detection.} In practical applications, short-video misinformation detection models trained on source domains may need to infer on unseen target domains, and the data distribution of these target domains may be different from source domains. We aim to propose specific modules to enhance the robust feature learning on the source domains $\mathcal{D}_s$ to further improve the model performance on the target domains $\mathcal{D}_t$.

\subsubsection{Overview of DOCTOR}

As illustrated in Figure \ref{fig:DOCTOR}, which demonstrates the domain generalization for the short-video misinformation detection, the proposed model, \textbf{DOCTOR}, is designed to address the generalization gap between source and target domains. 

The training process of the DOCTOR model is carried out in an end-to-end manner, consisting of two key modules that synchronize for optimization, jointly improving the model's domain generalization ability: 
(1) Cross-modal Interpolation Distillation: Construct interpolation features for video and audio modalities during the training process, and use them as teacher signals to guide student models to learn consistent and aligned representations between modalities. (2) Cross-modal Invariance Fusion: Construct multi-modal features and optimize them through a diffusion denoising mechanism, gradually enhancing the cross-domain invariance and discriminative power of the multi-modal representations.

The two modules are synchronously executed during the training process, jointly driving the model to learn robust representations from the perspectives of consistency and invariance, thereby improving generalization performance in the target domain.


\subsection{Cross-modal Interpolation Distillation}
In many domains, information falsification tends to be focused on distinct modalities (i.e., solely focusing on video or audio falsification). Enhancing domain generalization on them requires us to comprehensively integrate the inferential capabilities of each modality. However, in the cross-modal joint learning processes, multiple modalities may compete with each other, which can ultimately lead to insufficient knowledge acquisition in particular modalities and thus prevent optimal performance. To address this issue, we propose \textbf{Cross-modal Interpolation Distillation}.

Following the existing misinformation detection work \cite{bu2024fakingrecipe}, we utilize fine-tuned versions of HuBERT \cite{hsu2021hubert} as an encoder to extract audio features $X_a$, and Vision Transformer (ViT) \cite{dosovitskiy2020image} as an encoder to extract video features $X_v$. 

Due to the heterogeneity of modalities, the dimensions of the extracted features vary. Therefore, we employ the \textbf{Project} method to map the features of the video and audio modalities into a shared representation space, where all modalities have a unified dimension of $d$. In this shared representation space, we can directly establish connections between the different modalities. The unified dimensional feature representation we obtain is: $\hat{X}_k, k \in {a, v}$.
\begin{equation}
    \hat{X}_k = P_k (X_k) \in \mathcal{R}^d,
\end{equation}
where $P_k$ is the Project (different modal $k$ using different Project) and $k$ represents different modalities ($a: Audio, v: Video$). 

Then, in order to enhance the knowledge utilization of each modality in the cross-modal joint learning process and make the cross-modal representation more consistent in the shared representation space, we construct the interpolation representation between them through cross-modal representation mixup $\hat{X}_{a, v}$:
\begin{equation}
    \hat{X}_{a, v} = \theta \hat{X}_a + (1 - \theta) \hat{X}_v,
\end{equation}
where $\theta$ is a mixing ratio. 

If the loss of mixed representation $\hat{X}_{a, v}$ can be optimized to lower values, we can achieve the best state in all modalities at the same time. However, the amount of computation in multi-modal learning is often huge, and it is often difficult to achieve the optimal state, which is impractical in practical applications. Thus, we adopt the cross-modal distillation technique to distill the knowledge in the mixed representation into each modality and then optimize the learned representation to achieve the best state in all modalities. Specifically,
for each modal $k \in {(a: Audio, v: Video)}$, we initialize the student model $\gamma_t^k$ and utilize Simple Moving Average (SMA) to create the teacher model $\hat{\gamma_t^k}$, which is used to generate more stable and generalizable representations. The $\hat{\gamma}_{t}^{k}$ represents as:
\begin{equation}
\hat{\gamma}_{t}^{k} = 
\begin{cases} 
\gamma_{t}^{k}, & \text{if } t \leq t_0 \\
\frac{t - t_0}{t - t_0 + 1} \cdot \hat{\gamma}_{t-1}^{k} + \frac{1}{t - t_0 + 1} \cdot \gamma_{t}^{k}, & \text{if } t > t_0
\end{cases}
\end{equation}
where $\gamma_{t}^{k}$ means that iteration $t$ is the student mode state of modal $k$, and $t_0$ is the number of iterations at which to start the SMA. Therefore, the representation produced by the teacher model $\hat{\gamma}_{t}^{k}$ is denoted as $\dot{X}_k$, which serves as a stable reference for the modality $k$. We obtain a new cross-modal representation mixup $\dot{X}_{a, v}$ as follows:
\begin{equation}
    \dot{X}_{a, v} = \theta \dot{X}_a + (1 - \theta) \dot{X}_v.
\end{equation}

In the process of knowledge distillation and hybrid representation generation, in order to provide a flexible and controllable interpolation method when generating hybrid samples $\dot{X}_{a, v}$, we adopt Beta distribution. We define $\theta$ as $(\theta \sim \mathrm{Beta}(\alpha, \alpha))$ and $\alpha$ is a hyperparameter in the Beta distribution.

Due to the semantic gap between the modalities, we introduce a dynamic threshold $\theta$ to distinguish the dominance of different modalities and use the interpolated features of the modality closer to the $k$ modality ($k \in {(a: Audio, v: Video)}$) as the teacher signal to achieve cross-modal knowledge transfer. Therefore, the distillation loss should be specifically as follows:
\begin{equation}
\begin{split}
    L_{\text{dis}}^v = \left\| \hat{X}_v - \dot{X}_{a, v} \right\|_2^2, \quad \theta \leq 0.5 \\
    L_{\text{dis}}^a = \left\| \hat{X}_a - \dot{X}_{a, v} \right\|_2^2, \quad \theta > 0.5  
\end{split}
\end{equation}

Afterward, we assign a dedicated classifier to each modality before the \textit{Project} in the student model and optimize the features by the classification loss $L_{cls}^a$ and $L_{cls}^v$.

In addition, to bridge the gap between audio and video modalities and make the representations of each modality more consistent and synchronized, we further employ multi-modal supervised contrastive learning on a shared representation space. The contrastive learning loss formula is as follows:
\begin{equation}
    L_{con} = -\frac{1}{|B|} \sum_{i \in B} \log \frac{
        \exp(\hat{X}_a^i \cdot \hat{X}_v^i / \tau)
    }{
        \sum\limits_{\substack{j \in B \\ j \neq i}} \left[
            \exp(\hat{X}_a^i \cdot \hat{X}_a^j / \tau) + 
            \exp(\hat{X}_a^i \cdot \hat{X}_v^j / \tau)
        \right]
    },
\end{equation}
where $B$ is the sample set in the training batch, $\tau$ is the temperature coefficient, $\exp(\cdot / \tau)$ is a similarity scaling function used to adjust the discrimination between positive and negative samples, and $\hat{X}_a^i, \hat{X}_v^i$ are the audio and video embedding vectors of sample $i$.

Finally, we combine the losses of the above steps to obtain the final Cross-modal Interpolation Distillation loss: $Loss_{dis}$.
\begin{equation}
    Loss_{dis} = L_{{con}} + L_{cls}^a + L_{cls}^v + L_{dis}^a + L_{dis}^v.
\end{equation}


\subsection{Cross-modal Invariance Fusion}
When multi-modal joint fraud is emphasized in many domains, the separate process of each modality may not be adequate to enhance model generalization across these domains. Therefore, we need to comprehensively analyze cross-modal fusion features. However, direct fusion cross-modal features will accumulate the domain bias of each modality. In particular, for the video modality, domain biases present in numerous non-core video frames and non-core frame areas involved may accumulate significantly through the fusion of frames and areas on a frame-by-frame and area-by-area basis.

\begin{algorithm}
\caption{\textbf{Spatial Feature Extraction}}
\begin{algorithmic}[1]
\REQUIRE Frame $F_d$, Text Area ($TA$)
\ENSURE Spatial feature representation $X_{spa}$

\STATE \textbf{Text Area Detection and Embedding:}
\STATE Apply CRAFT OCR to detect $TA$ in frame $F_d$
\STATE Convert $TA$ to prompt embeddings $X_{vt}^d$ using SAM (Segment Anything Model) prompt encode

\STATE \textbf{Initial Visual Encoding:}
\STATE Input $F_d$ into pre-trained ViT to generate initial encoding $X_{v}^d$

\STATE \textbf{Bidirectional Attention Block:}
\STATE \textit{Self-Attention for Text Features:}
\STATE $\hat{X}_{vt}^d = \text{NORM}(X_{vt}^d + \text{Self-Att}(X_{vt}^d))$
\STATE \textit{Cross-Attention between Text and Visual Features:}
\STATE $\dot{X}_{vt}^d = \text{NORM}(\hat{X}_{vt}^d + \text{Cross-Att}(\hat{X}_{vt}^d, X_{v}^d))$
\STATE \textit{Cross-Attention for Enhanced Visual Features:}
\STATE $\hat{X}_{v}^d = \text{NORM}(X_{v}^d + \text{Cross-Att}(\text{MLP}(\dot{X}_{vt}^d), X_{v}^d))$

\STATE \textbf{Spatial Feature Extraction:}
\STATE Downsample $\hat{X}_{v}^d$ through two convolutional layers
\STATE Expand to obtain single-frame spatial pattern features $X^d$

\STATE \textbf{Multi-Frame Feature Fusion:}
\STATE Fuse all frame features: $\text{Self-Att}([X^1, X^2, \ldots, X^d])$
\STATE Average pooling: $X_{spa} = \text{MEAN}(\text{Self-Att}([X^1, X^2, \ldots, X^d]))$

\RETURN $X_{spa}$
\end{algorithmic}
\label{al:1}
\end{algorithm}

Toward this issue, we propose \textbf{Cross-modal Invariance Fusion}. It designs a novel mechanism for performing core extraction for video features from spatial and temporal perspectives. With extracted multi-modal features, we add noise to them step by step based on the diffusion model to filter out redundant details, thereby underlining the core features of multi-modal. Subsequently, by gradually mitigating the noise influence, we further outline the core information of multi-modal features through cross-modal mutual guidance, enhancing domain invariant features. We aim to mitigate the impact of domain bias and retain domain invariance to enhance the domain generalization ability for misinformation detection.

\subsubsection{Multimodal Feature Fusion}


For domain invariance of the \textit{\textbf{spatial features}} in the video, we consider that creators of misinformation often add misleading elements, such as subtitles, to disguise forged frames. Thus, we select text-rich frames ($F_d, d \in D$, where $D$ represents the set of all text-rich frames) as starting points. The extraction process for spatial features is detailed in Algorithm \ref{al:1}. 
Following the steps below, we obtain the spatial features $X_{spa}$.

To capture \textit{\textbf{temporal features}} in multi-modal content, we segment videos using text-rich frames $F_d$ as semantic anchors, as they typically encapsulate preceding context and exploit cognitive biases in visual memory—crucial for surfacing latent misinformation.

We represent each modality as a sequence of segments:
$Text = {(T_1, I_1), (T_2,  I_2) \dots, (T_n, I_n)}$,
$Video = {(V_1, I_1), \dots, (V_n, I_n)}$,
$Audio = {(A_1, I_1), \dots, (A_n, I_n)}$,
where $T_k$ denotes the $k$-th text segment (subtitle), $V_k$ and $A_k$ are the corresponding video frame and audio clip at interval $I_k$. The rate (fps) and total frames (vframes) are included to represent duration.
We first model the temporal structure of each modality independently, followed by a cross-modal interaction phase to capture joint temporal editing patterns.
For each time-aligned segment, we extract modality-specific features: text subtitles are encoded using BERT \cite{devlin2019bert} to obtain $Seg_t^i$, while video and audio features ($Seg_v^i$, $Seg_a^i$) are fused via self-attention.

\begin{table*}[h]
    \centering
    \small
    \caption{Performance comparison of domain generalization accuracy of different methods with DOCTOR on FakeSV and FakeTT datasets, and perform an ablation study on two key components (Section 3.2: Distillation and Section 3.3: Diffusion) of DOCTOR in domain generalization. * represents the domain generalization method.}
     \resizebox{0.9\textwidth}{!}{%
    \begin{tabular}{lcccccccccc}
        \toprule
        \textbf{Dataset} & \textbf{Method} & \textbf{Disaster} & \textbf{Society} & \textbf{Health} & \textbf{Culture} & \textbf{Politics} & \textbf{Science} & \textbf{Education} & \textbf{Finance} & \textbf{Military} \\
        \midrule
        \multirow{7}{*}{FakeSV} 
        & FakingRecipe   & 78.79 & 75.85 & 81.44 & 85.16 & 65.08 & 85.27 & 73.26 & 73.11 & 87.30 \\
        & SV-FEND        & 76.36 & 73.96 & 80.13 & 82.69 & 65.11 & 82.76 & 70.45 & 72.67 & 86.96 \\
        & OpEvFake       & 79.54 & 74.76 & 81.43 & 85.34 & 75.74 & 87.05 & 77.54 & 80.12 & 90.57 \\ 
         & CMRF*           & 79.76 & 74.45 & 81.28 & 84.77 & 74.49 & 86.56 & 76.32 & 80.76 & 91.87 \\
          & MDFEND*      &   61.39& 51.13 & 69.87 & 78.36 & 55.28&  79.74 & 71.33 & 59.26 & 66.65 \\
          & MMDFND*      &   75.49& 73.10 & 78.97 & 82.57 & 72.58&  82.75 & 74.28 & 79.13 & 86.67 \\
        
        & \textbf{DOCTOR} & \textbf{81.67} & \textbf{77.15} & \textbf{83.66} & \textbf{86.26} & \textbf{77.16} & \textbf{88.34} & \textbf{78.22} & \textbf{83.87} & \textbf{93.65} \\ 
        & w/o Diffusion  & 79.82 & 76.05 & 82.46 & 85.81 & 76.58 & 87.73 & 77.22 & 82.67 & 92.36 \\
        & w/o Distillation & 80.57 & 75.97 & 81.75 & 85.71 & 75.86 & 87.11 & 77.34 & 81.72 & 91.45 \\         
        \midrule
        \textbf{Dataset} & \textbf{Method} & \textbf{Disaster} & \textbf{Society} & \textbf{Health} & \textbf{Culture} & \textbf{Politics} & \textbf{Science} & \multicolumn{3}{c}{\textbf{\{Education, Finance, Military\}}}  \\
        \midrule
        \multirow{7}{*}{FakeTT} 
        & FakingRecipe   & 71.37 & 71.21 & 76.03 & 79.62 & 55.66 & 76.75 & \multicolumn{3}{c}{77.78} \\
        & SV-FEND        & 70.21 & 70.85 & 76.53 & 77.79 & 57.34 & 75.15 & \multicolumn{3}{c}{72.89} \\
        & OpEvFake       & 71.84 & 76.67 & 84.21 & 83.56 & 66.89 & 77.50 & \multicolumn{3}{c}{79.98} \\
         & CMRF*           & 71.75 & 77.05 & 81.87 & 83.47 & 65.78 & 74.96 & \multicolumn{3}{c}{81.43} \\
          & MDFEND*      &   62.23& 66.58 & 52.73 & 78.96 & 59.66&  70.13 &  \multicolumn{3}{c}{52.93} \\
          & MMDFND*      &    70.08 & 75.57 & 76.89 & 82.46 &63.44& 74.90&  \multicolumn{3}{c}{71.85}  \\
        
        & \textbf{DOCTOR}           & \textbf{74.18} & \textbf{78.34} & \textbf{85.33} & \textbf{88.31} & \textbf{70.33} & \textbf{78.23} & \multicolumn{3}{c}{\textbf{86.11}} \\
        & w/o Diffusion  & 73.52 & 77.97 & 84.47 & 85.18 & 67.83 & 76.38 & \multicolumn{3}{c}{85.18} \\
        & w/o Distillation & 72.45 & 77.44 & 83.77 & 84.61 & 68.95 & 77.85 & \multicolumn{3}{c}{81.72} \\
        \bottomrule
    \end{tabular}
    }
    \label{tab:1}
\end{table*}

To further capture temporal dynamics, we introduce two encoding strategies: Temporal Position Encoding (TPE) and Duration Encoding (DE). TPE uses equal-frequency binning to assign learnable embeddings based on both absolute and relative segment durations. DE employs sinusoidal functions to represent segment positions within the video. These obtain $TPE^i$ and $DE^i$ of the segment $i$.
The final segment representation is constructed as:
   $ Seg^i ( DE^i + TPE^i + Seg_t^i + Seg_v^i + Seg_a^i). $
Lastly, all segments are then fused via self-attention (Self-Att) to obtain the temporal representation $X_{time}$:
\begin{equation}
    X_{time} = \text{MEAN}(\text{Self-Att}([Seg^1,Seg^2,...,Seg^i]).
\end{equation}

\subsubsection{Diffusion Model Filtering Bias}



After obtaining the temporal feature $X_{time}$ and the spatial feature $X_{spa}$, we feed them into a diffusion model that progressively injects noise to filter out domain biases and highlight multi-modal features’ core information; as the noise influence gradually diminishes, we leverage cross‑modal interactions to further accentuate the fundamental information of misinformation forgery, boosting domain invariant features.

To more effectively capture the domain invariant features, we incorporate semantic and emotional features during the diffusion process, as they provide intent interpretability and contextual awareness for spatiotemporal features. Based on the analysis of previous studies \cite{bu2024fakingrecipe}, different modalities play a dominant role in conveying information: audio mainly expresses emotions, text (News Introduction and Title) conveys semantic information while also presenting emotional colors, and video usually supplements text to convey semantic content. We utilize the BERT to encode text, HuBERT to encode audio, and ViT to encode video to separately get $X_t$, $X_a$, $X_v$. The attention mechanism is used to obtain the emotional features $X_{emo}$ via $X_t$ and $X_a$, and the semantic features $X_{sem}$ via $X_t$ and $X_v$.

Based on the above multi-modal features, the diffusion model we designed has the following steps:

\textbf{In the \textit{forward diffusion} process}, we denote the initial distribution of each fusion feature as $X_{emo}^0$ (emotion feature), $X_{sem}^0$ (semantic feature), $X_{spa}^0$ (spatial feature), $X_{time}^0$ (temporal feature). For each feature, we gradually incorporate noise sampled from a Gaussian distribution into the true distribution. The steps are as:
\begin{equation}
    q(X_{F}^k \mid X_{F}^{k-1}) = \mathcal{N}(X_{F}^{k-1}; X_{F}^k\sqrt{1-\beta_F^k};I\beta_F^k),
\end{equation}
where the equation follows the Markov chain principle and represents the distribution of $X_{F}^k$ under $X_{F}^{k-1}$, $\beta_F^k$ is a weight and indicates the proportion of noise added at the $k$-th step. $I$ represents the identity matrix, which is used to control the covariance structure of the noise. $F$ is a feature of emo feature, sem feature, spa feature, and time feature. $X_{F}^k$ can be sampled through the following process:
\begin{equation}
    X_{F}^k =  x_{F}^0\sqrt{\prod_{i=1}^{k} \alpha_{F}^i} + \epsilon\sqrt{1 - \prod_{i=1}^{k} \alpha_{F}^i}, 
    \quad \alpha_{F}^i = 1 - \beta_{F}^i,
\end{equation}
where \(\epsilon\) is the noise sampled from \(\mathcal{N}(0, I)\).
The overall noise injection process is as follows, filtering out redundant biases:
\begin{equation}
    q(X_F^{1:K} \mid X_F^0) = \prod_{k=1}^{K} q(X_F^k \mid X_F^{k-1}).
\end{equation}

\textbf{In the \textit{backward diffusion} process}, we further enhance each basic feature by gradually alleviating the effect of noise and leveraging cross-modal features to guide each other. Specifically, we first fuse $X_{emo}^0$, $X_{sem}^0$, $X_{spa}^0$, $X_{time}^0$ into a mixed multi-features representation $X_{multi}^0$ through a full connect layer denoted by $\mathcal{F}_{fc}$.
\begin{equation}
    X_{multi}^0 = \mathcal{F}_{fc}(X_{emo}^0 \oplus X_{sem}^0 \oplus X_{spa}^0 \oplus X_{time}^0).
\end{equation}

\begin{table*}[h]
    \centering
    \caption{Comparison of general domain misinformation detection performance on FakeSV and FakeTT datasets.}
    \resizebox{0.85\textwidth}{!}{%
    \begin{tabular}{lccccccccc}
        \toprule
        \textbf{Dataset} & \textbf{Method} & \textbf{Acc.} & \textbf{F1} & \multicolumn{3}{c}{\textbf{Fake Class}} & \multicolumn{3}{c}{\textbf{Real Class}} \\
        \cmidrule(lr){5-7} \cmidrule(lr){8-10}
        & & & & \textbf{Precision} & \textbf{Recall} & \textbf{F1} & \textbf{Precision} & \textbf{Recall} & \textbf{F1} \\
        \midrule
        \multirow{10}{*}{FakeSV} 
        & FakingRecipe   & 82.83 & 82.62 & 85.28 & 83.88 & 84.58 & 79.84 & 81.51 & 80.67 \\
        & GPT-4         & 67.43 & 67.34 & 83.71 & 53.99 & 65.64 & 57.81 & 85.71 & 69.05 \\
        & GPT-4V        & 69.15 & 69.14 & 82.35 & 58.78 & 68.60 & 60.00 & 83.08 & 69.68 \\
        & FANVM         & 78.52 & 78.81 & 78.64 & 87.17 & 82.68 & 80.95 & 69.75 & 74.94 \\
        & SV-FEND       & 80.88 & 80.54 &  85.82 & 77.63 & 81.52 & 74.53 & 83.61 & 78.81 \\
        & TikTec        & 73.41 & 73.26 & 78.37 & 72.70 & 75.43 & 68.08 & 74.37 & 71.08 \\
        & HCFC-Medina   & 76.38 & 75.83 & 77.50 & 81.58 & 79.49 & 74.77 & 69.75 & 72.17 \\
        & HCFC-Hou      & 74.91 & 73.61 & 73.46 & 86.51 & 79.46 & 77.72 & 60.08 & 67.77 \\
        
        & \textbf{DOCTOR}          &  \textbf{84.91} &  \textbf{84.70} & \textbf{84.80} &  \textbf{91.57} &  \textbf{86.67} &  \textbf{83.46} & \textbf{82.57} &  \textbf{83.67} \\
        \midrule
        \multirow{9}{*}{FakeTT} 
        & FakingRecipe   & 78.58 & 76.62 &  65.49 & 74.75 & 69.81 & 86.56 &  80.50 & 83.42 \\
        & GPT-4         & 61.45 & 60.66 & 43.36 & 75.61 & 55.11 & 83.19 & 55.00 & 66.22 \\
        & GPT-4V        & 58.69 & 58.69 & 44.52 &  88.46 & 59.23 & 88.00 & 43.42 & 58.15 \\
        & FANVM         & 71.57 & 70.21 & 55.15 & 75.76 & 63.83 & 85.28 & 69.50 & 76.58 \\
        & SV-FEND       & 77.14 & 75.63 & 62.33 & 78.79 & 69.57 & 87.91 & 76.33 & 81.69 \\
        & TikTec        & 66.22 & 65.08 & 49.32 & 72.73 & 58.78 & 82.35 & 63.00 & 71.39 \\
        & HCFC-Medina   & 62.54 & 62.23 & 46.24 & 80.81 & 58.82 & 84.92 & 53.50 & 65.64 \\
        & HCFC-Hou      & 73.24 & 72.00 & 56.93 & 78.79 & 66.10 & 87.04 & 70.50 & 77.90 \\
        
        & \textbf{DOCTOR}          &  \textbf{78.92} &  \textbf{78.04} & \textbf{64.71} & \textbf{76.75} &  \textbf{72.21} &  \textbf{90.19} & \textbf{78.00} & \textbf{81.89} \\
        \bottomrule
    \end{tabular}}
    \label{tab:2}
\end{table*}

Then, the noise injection feature representation $X_{emo}^k$, $X_{sem}^k$, $X_{spa}^k$, $X_{time}^k$ obtained in the first step is combined with the fused multi-features representation $X_{multi}^0$ to gradually predict the distribution of each feature representation in the denoising model. In this way, each multi-modal feature, based on the information of other features, guides each other to restore the core features (the invariance characteristics of different domains misinformation forgery) of each modality to the greatest extent, so as to reduce the interference of domain bias. The specific steps are as follows:
\begin{equation}
    p(X_F^{k-1} \mid X_F^{k}) = \mathcal{N}(X_F^{k-1};\mu_\theta(X_F^{k},X_{multi}^{0},k),\Sigma_{\theta}(k)),
\end{equation}
where $\mu_\theta$ is the predicted mean (the direction of denoising), and $\Sigma_{\theta}(k)$ is the covariance of the noise. This process gradually generates $X_F^{k-1}$, $X_F^{k-2}$, ... , $X_F^{0}$ from $X_F^{k}$ that is, gradual denoising.

In total, under the optimization of the multi-feature cross-modal guided diffusion model, we can minimize the domain bias of misinformation domain generalization to retain the invariant of misinformation in different domains in the features.

Finally, we separately input the optimization features $\hat{X}_{emo}^0$ and $\hat{X}_{sem}^0$, $\hat{X}_{spa}^0$ and $\hat{X}_{time}^0$ into the MLP for prediction.

\section{Experiments}





\subsection{Experimental Setups}

\subsubsection{Datasets}

\textbf{(1) FakeSV} \cite{qi2023fakesv}: This dataset is the largest public Chinese video, audio, and text misinformation detection dataset, containing samples from two popular Chinese short-video platforms, TikTok and Kuaishou, covering news in multiple domains. 
\textbf{(2) FakeTT} \cite{bu2024fakingrecipe}: This dataset is a public English video, audio, and text misinformation detection dataset that focuses on videos related to events reported by the fact-checking website Snopes. 

We use Qwen2.5-72B \cite{yang2024qwen2} combined with manual verification for classification. We divide into nine domains (Society, Health, Disaster, Culture, Education, Finance, Politics, Science, and Military).

\subsubsection{Baselines}

We selected some popular misinformation short-video detection methods, such as FakingRecipe \cite{bu2024fakingrecipe}, OpEvFake \cite{zong2024unveiling}, TikTec \cite{shang2021multimodal}, HCFC-Hou \cite{hou2019towards}, HCFC-Medina \cite{serrano2020nlp}, FANVM \cite{choi2021using}, SV-FEND \cite{qi2023fakesv}. We also compared with the novel \textbf{general domain generalization} method CMRF \cite{fan2024cross} and the \textbf{text-image misinformation domain generalization} methods MMDFND \cite{tong2024mmdfnd} and MDFEND \cite{nan2021mdfend}. We also select large language models, such as GPT-4 \cite{achiam2023gpt} and GPT-4V \cite{yang2023dawn} for comparison.


\subsection{Experimental Results}

\subsubsection{Domain generalization experiments}

To evaluate DOCTOR in domain generalization, we show in Table \ref{tab:1} that DOCTOR consistently outperforms other methods across all domains.

Specifically, compared with the recent SOTA method for general domain generalization in video and audio, CMRF, our proposed DOCTOR model demonstrates a superior ability to effectively leverage multi-modal information while mitigating domain bias, thereby enhancing domain generalization in misinformation detection. As there is currently a lack of domain generalization methods tailored for short-video misinformation detection, we additionally compare the DOCTOR model with MMDFEND and MDFEND, which are SOTA methods (Open Source) for domain generalization in text-image misinformation detection. Across various domains, DOCTOR consistently achieves significantly higher accuracy.

Also, we compare with the recent SOTA short-video misinformation detection methods (such as FakingRecipe, SV-FEND, and OpEvFake). DOCTOR surpasses other models in terms of accuracy in domain generalization tests in the FakeSV and FakeTT datasets, reaching an average of 3\% \textasciitilde 4\% and effectively demonstrating the domain generalization ability of the DOCTOR model.

Finally, we conducted an ablation study on two key components of DOCTOR, Diffusion and Distillation, and found that the accuracy rate decreased. This indicates that these two components have an indispensable and effective synergy in the overall DOCTOR. 


\subsubsection{General domain experiments}


As shown in Table \ref{tab:2}, another point worth discussing is whether solving the differences between different domains will interfere with misinformation detection in the general domain. 
\textbf{(1)}
In the FakeSV dataset, DOCTOR achieved 84.91\% accuracy and 84.70\% F1. The overall performance is better than most existing short-video misinformation detection methods and some large language models. From the perspective of truth and misinformation, the DOCTOR is more effective in improving the method of identifying truth (Precision: 83.46\%, Recall: 82.57\%, F1: 83.67\%). DOCTOR is also very competitive in identifying misinformation methods;
\textbf{(2)}
In the FakeTT dataset, DOCTOR also has great advantages, achieving 78.92\% of accuracy and 78.04\% of F1. This result surpasses other methods. Similarly, as mentioned above, there are similar performance trends in the real class and fake class.

\subsubsection{Ablation experiments}

To evaluate the effectiveness of each component in the DOCTOR model, we perform comparative analyses by omitting each component individually. The experimental setups are as follows:
(1) w/o Diffusion: The diffusion model module is omitted, and fused features are used directly.
(2) w/o SE: Semantic and emotion feature fusion is removed.
(3) w/o ST: Temporal and spatial feature fusion is removed.
(4) w/o Distillation: The interpolation knowledge distillation module is omitted.

As detailed in Table \ref{tab:3}, we present the experimental results of the ablation studies. We could observe that the original DOCTOR outperforms all variants, demonstrating
the efficacy of each component. Overall, our findings lead to the following conclusions: (1) The importance of spatiotemporal features is usually higher than that of emotional semantic features, which reflects the uniqueness of the spatiotemporal relationship of video angles. (2) The improvement of the diffusion model is slightly better than that of the interpolation distillation, but the difference is not significant, which reflects the common effectiveness of the two.
\begin{table}[h]
    \centering
    \caption{Ablation Study of DOCTOR in the general domain.}
    \resizebox{\linewidth}{!}{%
    \begin{tabular}{lcccccc}
        \toprule
        \textbf{Dataset} & \textbf{Method} & \textbf{F1} & \textbf{Acc.} & \textbf{AUC} & \textbf{Recall} & \textbf{Precision} \\
        \midrule
        
        \multirow{5}{*}{FakeSV} 
          & \textbf{DOCTOR}             & \textbf{84.70}  & \textbf{84.91} & \textbf{83.82} & \textbf{83.82} & \textbf{84.62} \\
          & w/o Diffusion    & 82.54  & 82.84  & 82.46  & 82.46  & 82.62 \\
          & w/o SE           & 83.76  & 83.67  & 82.76  & 82.76  & 82.98 \\
          & w/o ST           & 82.97  & 83.42  & 82.34  & 82.34  & 82.67 \\
          & w/o Distillation & 82.90  & 83.21  & 82.79  & 82.79  & 83.03 \\
        \midrule
        
        \multirow{5}{*}{FakeTT} 
          & \textbf{DOCTOR}             & \textbf{78.04}  & \textbf{78.92} & \textbf{77.37} & \textbf{77.37} & \textbf{76.44} \\
          & w/o Diffusion    & 74.79  & 76.25  & 76.89  & 76.89  & 74.32 \\
          & w/o SE           & 76.63  & 77.54  & 76.87  & 76.87  & 74.54 \\
          & w/o ST           & 74.82  & 74.23  & 75.47  & 75.47  & 72.67 \\
          & w/o Distillation & 76.44  & 77.92  & 76.87  & 76.87  & 75.85 \\
        \bottomrule
    \end{tabular}}
    \label{tab:3}
\end{table}

\begin{table}[h]
\centering
\caption{Learning rate for DOCTOR model (Accuracy).}
\resizebox{\linewidth}{!}{%
\begin{tabular}{l|c|c|c|c|c|c}
\hline
\textbf{Dataset} &  \textbf{1e-5}  & \textbf{5e-5}  & \textbf{1e-4}  & \textbf{5e-4}  & \textbf{1e-3}  & \textbf{5e-3}  \\
\hline
\textbf{FakeSV}           &  82.32  &  \textbf{84.91}  & 83.45  & 80.78  & 79.53  & 78.44  \\ 
\midrule
\textbf{FakeTT}          &  72.24  & 74.31  & 76.86 & 78.57  &  \textbf{78.92}  & 77.39  \\
\hline
\end{tabular}}
\label{tab:lr_results}
\end{table}

\subsubsection{Parameter sensitivity analysis}
We systematically evaluate DOCTOR's performance across the learning rate. The results are summarized in Table \ref{tab:lr_results}. From the analysis, we observe that the optimal learning rate for DOCTOR varies depending on the dataset: 5e-5 achieves the best performance on the FakeSV dataset, while 1e-3 yields the highest performance on the FakeTT dataset.



\subsubsection{Case study}

\begin{figure}
    \centering
    \includegraphics[width=\linewidth]{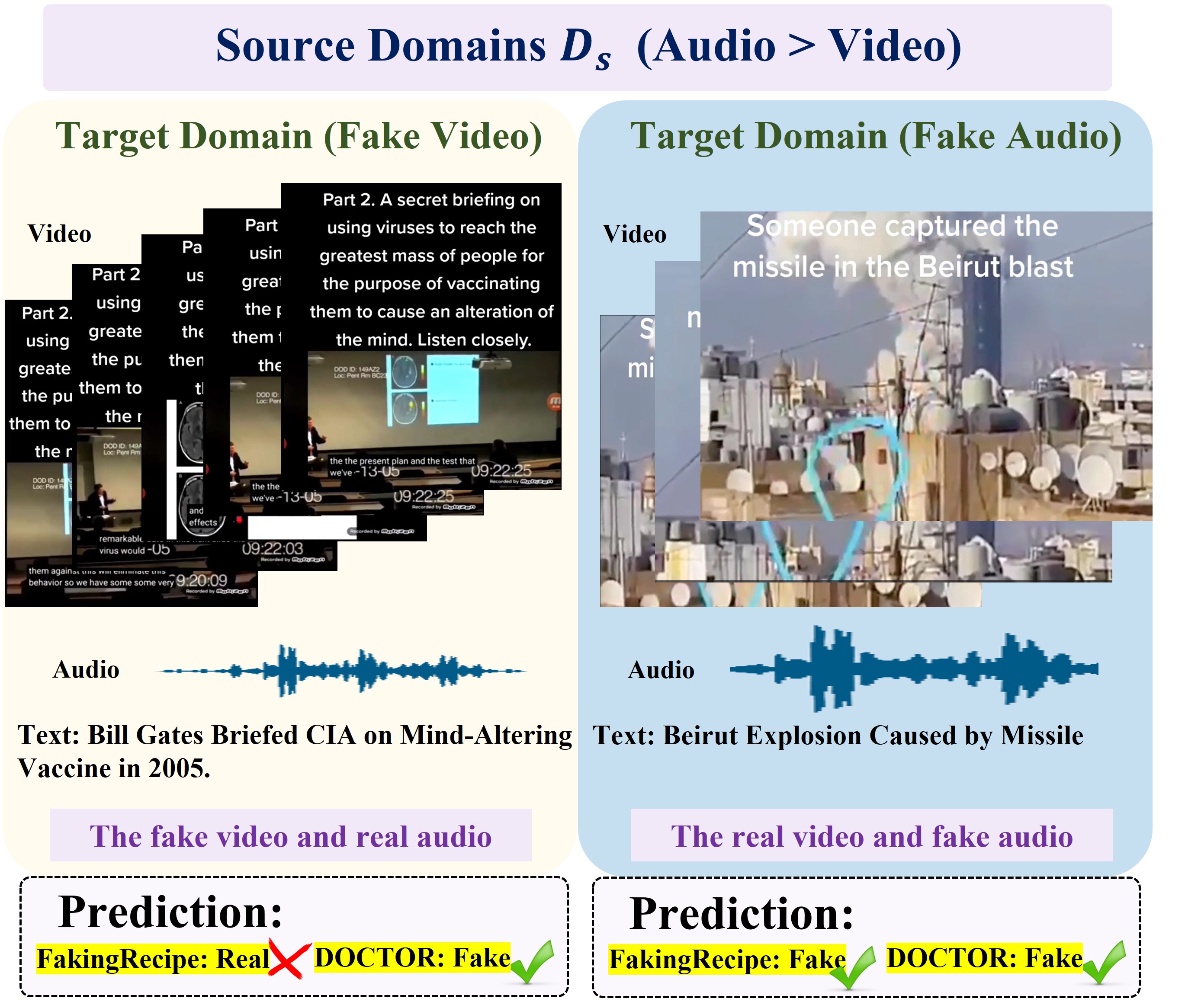}
    \caption{Case Study illustrating the effectiveness of DOCTOR model in cross-modal forgery detection.}
    \label{fig:case}
\end{figure}

As shown in Figure \ref{fig:case}, when the model is trained with a focus on audio forgery detection in the source domain, our case study demonstrates that DOCTOR achieves superior performance in detecting video forgeries in the target domain. This improvement is largely attributed to the use of cross-modal interpolation distillation, which facilitates effective knowledge transfer across modalities and enhances the model’s generalization ability and robustness in cross-modal forgery detection tasks.

\section{Conclusion}

In this paper, we firstly explore the shortcomings of the domain generalization of short-video misinformation detection, which are mainly divided into two aspects: \textbf{(1) Varying Modal-dependence in Different Domains} and \textbf{(2) Domain-bias Enhancement in Cross-modal Fusion}. To bridge these gaps, we propose an innovative DOmain generalization model via ConsisTency and invariance learning for shORt-video misinformation detection (named \textbf{DOCTOR}). It incorporates two core modules: \textbf{(a) Cross-modal Interpolation Distillation} and \textbf{(b) Cross-modal Invariance Fusion}. We validate the domain generalization effectiveness of misinformation detection of DOCTOR through comprehensive experiments. In addition, in-depth analysis, including ablation studies, demonstrates the design rationality of each module in the DOCTOR model. 

\begin{acks}

This work was supported by the National Key R\&D Program of China under Grant No. 2022YFC3303600, the Zhejiang Provincial Natural Science Foundation of China under Grant No. LY23F020010, and the National Natural Science Foundation of China under Grant No. 62337001.

\end{acks}

\bibliography{abbrev}


\begin{thebibliography}{49}


\ifx \showCODEN    \undefined \def \showCODEN     #1{\unskip}     \fi
\ifx \showISBNx    \undefined \def \showISBNx     #1{\unskip}     \fi
\ifx \showISBNxiii \undefined \def \showISBNxiii  #1{\unskip}     \fi
\ifx \showISSN     \undefined \def \showISSN      #1{\unskip}     \fi
\ifx \showLCCN     \undefined \def \showLCCN      #1{\unskip}     \fi
\ifx \shownote     \undefined \def \shownote      #1{#1}          \fi
\ifx \showarticletitle \undefined \def \showarticletitle #1{#1}   \fi
\ifx \showURL      \undefined \def \showURL       {\relax}        \fi
\providecommand\bibfield[2]{#2}
\providecommand\bibinfo[2]{#2}
\providecommand\natexlab[1]{#1}
\providecommand\showeprint[2][]{arXiv:#2}

\bibitem[Achiam et~al\mbox{.}(2023)]%
        {achiam2023gpt}
\bibfield{author}{\bibinfo{person}{Josh Achiam}, \bibinfo{person}{Steven Adler}, \bibinfo{person}{Sandhini Agarwal}, \bibinfo{person}{Lama Ahmad}, \bibinfo{person}{Ilge Akkaya}, \bibinfo{person}{Florencia~Leoni Aleman}, \bibinfo{person}{Diogo Almeida}, \bibinfo{person}{Janko Altenschmidt}, \bibinfo{person}{Sam Altman}, \bibinfo{person}{Shyamal Anadkat}, {et~al\mbox{.}}} \bibinfo{year}{2023}\natexlab{}.
\newblock \showarticletitle{Gpt-4 technical report}.
\newblock \bibinfo{journal}{\emph{arXiv preprint arXiv:2303.08774}} (\bibinfo{year}{2023}).
\newblock


\bibitem[Agarwal et~al\mbox{.}(2024)]%
        {agarwal2024television}
\bibfield{author}{\bibinfo{person}{Anmol Agarwal}, \bibinfo{person}{Pratyush Priyadarshi}, \bibinfo{person}{Shiven Sinha}, \bibinfo{person}{Shrey Gupta}, \bibinfo{person}{Hitkul Jangra}, \bibinfo{person}{Ponnurangam Kumaraguru}, {and} \bibinfo{person}{Kiran Garimella}.} \bibinfo{year}{2024}\natexlab{}.
\newblock \showarticletitle{Television discourse decoded: Comprehensive multimodal analytics at scale}. In \bibinfo{booktitle}{\emph{Proceedings of the 30th ACM SIGKDD Conference on Knowledge Discovery and Data Mining}}. \bibinfo{pages}{4752--4763}.
\newblock


\bibitem[Ali et~al\mbox{.}(2022)]%
        {ali2022effects}
\bibfield{author}{\bibinfo{person}{Khudejah Ali}, \bibinfo{person}{Cong Li}, \bibinfo{person}{Syed~Ali Muqtadir}, {et~al\mbox{.}}} \bibinfo{year}{2022}\natexlab{}.
\newblock \showarticletitle{The effects of emotions, individual attitudes towards vaccination, and social endorsements on perceived fake news credibility and sharing motivations}.
\newblock \bibinfo{journal}{\emph{Computers in Human Behavior}}  \bibinfo{volume}{134} (\bibinfo{year}{2022}), \bibinfo{pages}{107307}.
\newblock


\bibitem[Bu et~al\mbox{.}(2023)]%
        {bu2023combating}
\bibfield{author}{\bibinfo{person}{Yuyan Bu}, \bibinfo{person}{Qiang Sheng}, \bibinfo{person}{Juan Cao}, \bibinfo{person}{Peng Qi}, \bibinfo{person}{Danding Wang}, {and} \bibinfo{person}{Jintao Li}.} \bibinfo{year}{2023}\natexlab{}.
\newblock \showarticletitle{Combating online misinformation videos: Characterization, detection, and future directions}. In \bibinfo{booktitle}{\emph{Proceedings of the 31st ACM International Conference on Multimedia}}. \bibinfo{pages}{8770--8780}.
\newblock


\bibitem[Bu et~al\mbox{.}(2024)]%
        {bu2024fakingrecipe}
\bibfield{author}{\bibinfo{person}{Yuyan Bu}, \bibinfo{person}{Qiang Sheng}, \bibinfo{person}{Juan Cao}, \bibinfo{person}{Peng Qi}, \bibinfo{person}{Danding Wang}, {and} \bibinfo{person}{Jintao Li}.} \bibinfo{year}{2024}\natexlab{}.
\newblock \showarticletitle{Fakingrecipe: Detecting fake news on short video platforms from the perspective of creative process}. In \bibinfo{booktitle}{\emph{Proceedings of the 32nd ACM International Conference on Multimedia}}. \bibinfo{pages}{1351--1360}.
\newblock


\bibitem[Castelo et~al\mbox{.}(2019)]%
        {castelo2019topic}
\bibfield{author}{\bibinfo{person}{Sonia Castelo}, \bibinfo{person}{Thais Almeida}, \bibinfo{person}{Anas Elghafari}, \bibinfo{person}{A{\'e}cio Santos}, \bibinfo{person}{Kien Pham}, \bibinfo{person}{Eduardo Nakamura}, {and} \bibinfo{person}{Juliana Freire}.} \bibinfo{year}{2019}\natexlab{}.
\newblock \showarticletitle{A topic-agnostic approach for identifying fake news pages}. In \bibinfo{booktitle}{\emph{Companion proceedings of the 2019 World Wide Web conference}}. \bibinfo{pages}{975--980}.
\newblock


\bibitem[Chen et~al\mbox{.}(2022)]%
        {chen2022cross}
\bibfield{author}{\bibinfo{person}{Yixuan Chen}, \bibinfo{person}{Dongsheng Li}, \bibinfo{person}{Peng Zhang}, \bibinfo{person}{Jie Sui}, \bibinfo{person}{Qin Lv}, \bibinfo{person}{Lu Tun}, {and} \bibinfo{person}{Li Shang}.} \bibinfo{year}{2022}\natexlab{}.
\newblock \showarticletitle{Cross-modal ambiguity learning for multimodal fake news detection}. In \bibinfo{booktitle}{\emph{Proceedings of the ACM web conference 2022}}. \bibinfo{pages}{2897--2905}.
\newblock


\bibitem[Choi and Ko(2021)]%
        {choi2021using}
\bibfield{author}{\bibinfo{person}{Hyewon Choi} {and} \bibinfo{person}{Youngjoong Ko}.} \bibinfo{year}{2021}\natexlab{}.
\newblock \showarticletitle{Using topic modeling and adversarial neural networks for fake news video detection}. In \bibinfo{booktitle}{\emph{Proceedings of the 30th ACM international conference on information \& knowledge management}}. \bibinfo{pages}{2950--2954}.
\newblock


\bibitem[Devlin et~al\mbox{.}(2019)]%
        {devlin2019bert}
\bibfield{author}{\bibinfo{person}{Jacob Devlin}, \bibinfo{person}{Ming-Wei Chang}, \bibinfo{person}{Kenton Lee}, {and} \bibinfo{person}{Kristina Toutanova}.} \bibinfo{year}{2019}\natexlab{}.
\newblock \showarticletitle{Bert: Pre-training of deep bidirectional transformers for language understanding}. In \bibinfo{booktitle}{\emph{Proceedings of the 2019 conference of the North American chapter of the association for computational linguistics: human language technologies, volume 1 (long and short papers)}}. \bibinfo{pages}{4171--4186}.
\newblock


\bibitem[Dong et~al\mbox{.}(2023)]%
        {dong2023simmmdg}
\bibfield{author}{\bibinfo{person}{Hao Dong}, \bibinfo{person}{Ismail Nejjar}, \bibinfo{person}{Han Sun}, \bibinfo{person}{Eleni Chatzi}, {and} \bibinfo{person}{Olga Fink}.} \bibinfo{year}{2023}\natexlab{}.
\newblock \showarticletitle{SimMMDG: A simple and effective framework for multi-modal domain generalization}.
\newblock \bibinfo{journal}{\emph{Advances in Neural Information Processing Systems}}  \bibinfo{volume}{36} (\bibinfo{year}{2023}), \bibinfo{pages}{78674--78695}.
\newblock


\bibitem[Dosovitskiy et~al\mbox{.}(2020)]%
        {dosovitskiy2020image}
\bibfield{author}{\bibinfo{person}{Alexey Dosovitskiy}, \bibinfo{person}{Lucas Beyer}, \bibinfo{person}{Alexander Kolesnikov}, \bibinfo{person}{Dirk Weissenborn}, \bibinfo{person}{Xiaohua Zhai}, \bibinfo{person}{Thomas Unterthiner}, \bibinfo{person}{Mostafa Dehghani}, \bibinfo{person}{Matthias Minderer}, \bibinfo{person}{G Heigold}, \bibinfo{person}{S Gelly}, {et~al\mbox{.}}} \bibinfo{year}{2020}\natexlab{}.
\newblock \showarticletitle{An Image is Worth 16x16 Words: Transformers for Image Recognition at Scale}. In \bibinfo{booktitle}{\emph{International Conference on Learning Representations}}.
\newblock


\bibitem[Fan et~al\mbox{.}(2024)]%
        {fan2024cross}
\bibfield{author}{\bibinfo{person}{Yunfeng Fan}, \bibinfo{person}{Wenchao Xu}, \bibinfo{person}{Haozhao Wang}, {and} \bibinfo{person}{Song Guo}.} \bibinfo{year}{2024}\natexlab{}.
\newblock \showarticletitle{Cross-modal representation flattening for multi-modal domain generalization}.
\newblock \bibinfo{journal}{\emph{Advances in Neural Information Processing Systems}}  \bibinfo{volume}{37} (\bibinfo{year}{2024}), \bibinfo{pages}{66773--66795}.
\newblock


\bibitem[Hou et~al\mbox{.}(2019)]%
        {hou2019towards}
\bibfield{author}{\bibinfo{person}{Rui Hou}, \bibinfo{person}{Ver{\'o}nica P{\'e}rez-Rosas}, \bibinfo{person}{Stacy Loeb}, {and} \bibinfo{person}{Rada Mihalcea}.} \bibinfo{year}{2019}\natexlab{}.
\newblock \showarticletitle{Towards automatic detection of misinformation in online medical videos}. In \bibinfo{booktitle}{\emph{2019 International conference on multimodal interaction}}. \bibinfo{pages}{235--243}.
\newblock


\bibitem[Hsu et~al\mbox{.}(2021)]%
        {hsu2021hubert}
\bibfield{author}{\bibinfo{person}{Wei-Ning Hsu}, \bibinfo{person}{Benjamin Bolte}, \bibinfo{person}{Yao-Hung~Hubert Tsai}, \bibinfo{person}{Kushal Lakhotia}, \bibinfo{person}{Ruslan Salakhutdinov}, {and} \bibinfo{person}{Abdelrahman Mohamed}.} \bibinfo{year}{2021}\natexlab{}.
\newblock \showarticletitle{Hubert: Self-supervised speech representation learning by masked prediction of hidden units}.
\newblock \bibinfo{journal}{\emph{IEEE/ACM transactions on audio, speech, and language processing}}  \bibinfo{volume}{29} (\bibinfo{year}{2021}), \bibinfo{pages}{3451--3460}.
\newblock


\bibitem[Hu et~al\mbox{.}(2024)]%
        {hu2024cross}
\bibfield{author}{\bibinfo{person}{Xihang Hu}, \bibinfo{person}{Fuming Sun}, \bibinfo{person}{Jing Sun}, \bibinfo{person}{Fasheng Wang}, {and} \bibinfo{person}{Haojie Li}.} \bibinfo{year}{2024}\natexlab{}.
\newblock \showarticletitle{Cross-modal fusion and progressive decoding network for RGB-D salient object detection}.
\newblock \bibinfo{journal}{\emph{International Journal of Computer Vision}} \bibinfo{volume}{132}, \bibinfo{number}{8} (\bibinfo{year}{2024}), \bibinfo{pages}{3067--3085}.
\newblock


\bibitem[Jin et~al\mbox{.}(2022)]%
        {jin2022evaluating}
\bibfield{author}{\bibinfo{person}{Weina Jin}, \bibinfo{person}{Xiaoxiao Li}, {and} \bibinfo{person}{Ghassan Hamarneh}.} \bibinfo{year}{2022}\natexlab{}.
\newblock \showarticletitle{Evaluating explainable AI on a multi-modal medical imaging task: Can existing algorithms fulfill clinical requirements?}. In \bibinfo{booktitle}{\emph{Proceedings of the AAAI Conference on Artificial Intelligence}}, Vol.~\bibinfo{volume}{36}. \bibinfo{pages}{11945--11953}.
\newblock


\bibitem[Li et~al\mbox{.}(2021)]%
        {li2021entity}
\bibfield{author}{\bibinfo{person}{Peiguang Li}, \bibinfo{person}{Xian Sun}, \bibinfo{person}{Hongfeng Yu}, \bibinfo{person}{Yu Tian}, \bibinfo{person}{Fanglong Yao}, {and} \bibinfo{person}{Guangluan Xu}.} \bibinfo{year}{2021}\natexlab{}.
\newblock \showarticletitle{Entity-oriented multi-modal alignment and fusion network for fake news detection}.
\newblock \bibinfo{journal}{\emph{IEEE Transactions on Multimedia}}  \bibinfo{volume}{24} (\bibinfo{year}{2021}), \bibinfo{pages}{3455--3468}.
\newblock


\bibitem[Liu et~al\mbox{.}(2024a)]%
        {liu2024continual}
\bibfield{author}{\bibinfo{person}{RuiQi Liu}, \bibinfo{person}{Boyu Diao}, \bibinfo{person}{Libo Huang}, \bibinfo{person}{Zijia An}, \bibinfo{person}{Zhulin An}, {and} \bibinfo{person}{Yongjun Xu}.} \bibinfo{year}{2024}\natexlab{a}.
\newblock \showarticletitle{Continual learning in the frequency domain}.
\newblock \bibinfo{journal}{\emph{Advances in Neural Information Processing Systems}}  \bibinfo{volume}{37} (\bibinfo{year}{2024}), \bibinfo{pages}{85389--85411}.
\newblock


\bibitem[Liu et~al\mbox{.}(2024b)]%
        {liu2024fka}
\bibfield{author}{\bibinfo{person}{Xuannan Liu}, \bibinfo{person}{Peipei Li}, \bibinfo{person}{Huaibo Huang}, \bibinfo{person}{Zekun Li}, \bibinfo{person}{Xing Cui}, \bibinfo{person}{Jiahao Liang}, \bibinfo{person}{Lixiong Qin}, \bibinfo{person}{Weihong Deng}, {and} \bibinfo{person}{Zhaofeng He}.} \bibinfo{year}{2024}\natexlab{b}.
\newblock \showarticletitle{Fka-owl: Advancing multimodal fake news detection through knowledge-augmented lvlms}. In \bibinfo{booktitle}{\emph{Proceedings of the 32nd ACM International Conference on Multimedia}}. \bibinfo{pages}{10154--10163}.
\newblock


\bibitem[Ma et~al\mbox{.}(2024)]%
        {ma2024cross}
\bibfield{author}{\bibinfo{person}{Xinran Ma}, \bibinfo{person}{Mouxing Yang}, \bibinfo{person}{Yunfan Li}, \bibinfo{person}{Peng Hu}, \bibinfo{person}{Jiancheng Lv}, {and} \bibinfo{person}{Xi Peng}.} \bibinfo{year}{2024}\natexlab{}.
\newblock \showarticletitle{Cross-modal retrieval with noisy correspondence via consistency refining and mining}.
\newblock \bibinfo{journal}{\emph{IEEE transactions on image processing}} (\bibinfo{year}{2024}).
\newblock


\bibitem[Nan et~al\mbox{.}(2021)]%
        {nan2021mdfend}
\bibfield{author}{\bibinfo{person}{Qiong Nan}, \bibinfo{person}{Juan Cao}, \bibinfo{person}{Yongchun Zhu}, \bibinfo{person}{Yanyan Wang}, {and} \bibinfo{person}{Jintao Li}.} \bibinfo{year}{2021}\natexlab{}.
\newblock \showarticletitle{MDFEND: Multi-domain fake news detection}. In \bibinfo{booktitle}{\emph{Proceedings of the 30th ACM international conference on information \& knowledge management}}. \bibinfo{pages}{3343--3347}.
\newblock


\bibitem[Oh et~al\mbox{.}(2023)]%
        {oh2023geodesic}
\bibfield{author}{\bibinfo{person}{Changdae Oh}, \bibinfo{person}{Junhyuk So}, \bibinfo{person}{Hoyoon Byun}, \bibinfo{person}{YongTaek Lim}, \bibinfo{person}{Minchul Shin}, \bibinfo{person}{Jong-June Jeon}, {and} \bibinfo{person}{Kyungwoo Song}.} \bibinfo{year}{2023}\natexlab{}.
\newblock \showarticletitle{Geodesic multi-modal mixup for robust fine-tuning}.
\newblock \bibinfo{journal}{\emph{Advances in Neural Information Processing Systems}}  \bibinfo{volume}{36} (\bibinfo{year}{2023}), \bibinfo{pages}{52326--52341}.
\newblock


\bibitem[Ouali et~al\mbox{.}(2023)]%
        {ouali2023black}
\bibfield{author}{\bibinfo{person}{Yassine Ouali}, \bibinfo{person}{Adrian Bulat}, \bibinfo{person}{Brais Matinez}, {and} \bibinfo{person}{Georgios Tzimiropoulos}.} \bibinfo{year}{2023}\natexlab{}.
\newblock \showarticletitle{Black box few-shot adaptation for vision-language models}. In \bibinfo{booktitle}{\emph{Proceedings of the IEEE/CVF International Conference on Computer Vision}}. \bibinfo{pages}{15534--15546}.
\newblock


\bibitem[Pang et~al\mbox{.}(2022)]%
        {pang2022tackling}
\bibfield{author}{\bibinfo{person}{Hua Pang}, \bibinfo{person}{Jun Liu}, {and} \bibinfo{person}{Jiahui Lu}.} \bibinfo{year}{2022}\natexlab{}.
\newblock \showarticletitle{Tackling fake news in socially mediated public spheres: A comparison of Weibo and WeChat}.
\newblock \bibinfo{journal}{\emph{Technology in Society}}  \bibinfo{volume}{70} (\bibinfo{year}{2022}), \bibinfo{pages}{102004}.
\newblock


\bibitem[Planamente et~al\mbox{.}(2022)]%
        {planamente2022domain}
\bibfield{author}{\bibinfo{person}{Mirco Planamente}, \bibinfo{person}{Chiara Plizzari}, \bibinfo{person}{Emanuele Alberti}, {and} \bibinfo{person}{Barbara Caputo}.} \bibinfo{year}{2022}\natexlab{}.
\newblock \showarticletitle{Domain generalization through audio-visual relative norm alignment in first person action recognition}. In \bibinfo{booktitle}{\emph{Proceedings of the IEEE/CVF winter conference on applications of computer vision}}. \bibinfo{pages}{1807--1818}.
\newblock


\bibitem[Qi et~al\mbox{.}(2023)]%
        {qi2023fakesv}
\bibfield{author}{\bibinfo{person}{Peng Qi}, \bibinfo{person}{Yuyan Bu}, \bibinfo{person}{Juan Cao}, \bibinfo{person}{Wei Ji}, \bibinfo{person}{Ruihao Shui}, \bibinfo{person}{Junbin Xiao}, \bibinfo{person}{Danding Wang}, {and} \bibinfo{person}{Tat-Seng Chua}.} \bibinfo{year}{2023}\natexlab{}.
\newblock \showarticletitle{Fakesv: A multimodal benchmark with rich social context for fake news detection on short video platforms}. In \bibinfo{booktitle}{\emph{Proceedings of the AAAI Conference on Artificial Intelligence}}, Vol.~\bibinfo{volume}{37}. \bibinfo{pages}{14444--14452}.
\newblock


\bibitem[Qi et~al\mbox{.}(2021)]%
        {qi2021improving}
\bibfield{author}{\bibinfo{person}{Peng Qi}, \bibinfo{person}{Juan Cao}, \bibinfo{person}{Xirong Li}, \bibinfo{person}{Huan Liu}, \bibinfo{person}{Qiang Sheng}, \bibinfo{person}{Xiaoyue Mi}, \bibinfo{person}{Qin He}, \bibinfo{person}{Yongbiao Lv}, \bibinfo{person}{Chenyang Guo}, {and} \bibinfo{person}{Yingchao Yu}.} \bibinfo{year}{2021}\natexlab{}.
\newblock \showarticletitle{Improving fake news detection by using an entity-enhanced framework to fuse diverse multimodal clues}. In \bibinfo{booktitle}{\emph{Proceedings of the 29th ACM International Conference on Multimedia}}. \bibinfo{pages}{1212--1220}.
\newblock


\bibitem[Serrano et~al\mbox{.}(2020)]%
        {serrano2020nlp}
\bibfield{author}{\bibinfo{person}{Juan Carlos~Medina Serrano}, \bibinfo{person}{Orestis Papakyriakopoulos}, {and} \bibinfo{person}{Simon Hegelich}.} \bibinfo{year}{2020}\natexlab{}.
\newblock \showarticletitle{NLP-based feature extraction for the detection of COVID-19 misinformation videos on YouTube}. In \bibinfo{booktitle}{\emph{Proceedings of the 1st Workshop on NLP for COVID-19 at ACL 2020}}.
\newblock


\bibitem[Shang et~al\mbox{.}(2021)]%
        {shang2021multimodal}
\bibfield{author}{\bibinfo{person}{Lanyu Shang}, \bibinfo{person}{Ziyi Kou}, \bibinfo{person}{Yang Zhang}, {and} \bibinfo{person}{Dong Wang}.} \bibinfo{year}{2021}\natexlab{}.
\newblock \showarticletitle{A multimodal misinformation detector for covid-19 short videos on tiktok}. In \bibinfo{booktitle}{\emph{2021 IEEE international conference on big data (big data)}}. IEEE, \bibinfo{pages}{899--908}.
\newblock


\bibitem[Silva et~al\mbox{.}(2021)]%
        {silva2021embracing}
\bibfield{author}{\bibinfo{person}{Amila Silva}, \bibinfo{person}{Ling Luo}, \bibinfo{person}{Shanika Karunasekera}, {and} \bibinfo{person}{Christopher Leckie}.} \bibinfo{year}{2021}\natexlab{}.
\newblock \showarticletitle{Embracing domain differences in fake news: Cross-domain fake news detection using multi-modal data}. In \bibinfo{booktitle}{\emph{Proceedings of the AAAI conference on artificial intelligence}}, Vol.~\bibinfo{volume}{35}. \bibinfo{pages}{557--565}.
\newblock


\bibitem[Tong et~al\mbox{.}(2024)]%
        {tong2024mmdfnd}
\bibfield{author}{\bibinfo{person}{Yu Tong}, \bibinfo{person}{Weihai Lu}, \bibinfo{person}{Zhe Zhao}, \bibinfo{person}{Song Lai}, {and} \bibinfo{person}{Tong Shi}.} \bibinfo{year}{2024}\natexlab{}.
\newblock \showarticletitle{MMDFND: Multi-modal Multi-Domain Fake News Detection}. In \bibinfo{booktitle}{\emph{Proceedings of the 32nd ACM International Conference on Multimedia}}. \bibinfo{pages}{1178--1186}.
\newblock


\bibitem[Wang et~al\mbox{.}(2023)]%
        {wang2023cross}
\bibfield{author}{\bibinfo{person}{Longzheng Wang}, \bibinfo{person}{Chuang Zhang}, \bibinfo{person}{Hongbo Xu}, \bibinfo{person}{Yongxiu Xu}, \bibinfo{person}{Xiaohan Xu}, {and} \bibinfo{person}{Siqi Wang}.} \bibinfo{year}{2023}\natexlab{}.
\newblock \showarticletitle{Cross-modal contrastive learning for multimodal fake news detection}. In \bibinfo{booktitle}{\emph{Proceedings of the 31st ACM international conference on multimedia}}. \bibinfo{pages}{5696--5704}.
\newblock


\bibitem[Wang et~al\mbox{.}(2025)]%
        {wang2025cross}
\bibfield{author}{\bibinfo{person}{Tianshi Wang}, \bibinfo{person}{Fengling Li}, \bibinfo{person}{Lei Zhu}, \bibinfo{person}{Jingjing Li}, \bibinfo{person}{Zheng Zhang}, {and} \bibinfo{person}{Heng~Tao Shen}.} \bibinfo{year}{2025}\natexlab{}.
\newblock \showarticletitle{Cross-modal retrieval: a systematic review of methods and future directions}.
\newblock \bibinfo{journal}{\emph{Proc. IEEE}} (\bibinfo{year}{2025}).
\newblock


\bibitem[Wang et~al\mbox{.}(2024)]%
        {wang2024search}
\bibfield{author}{\bibinfo{person}{Zixin Wang}, \bibinfo{person}{Yadan Luo}, \bibinfo{person}{Liang Zheng}, \bibinfo{person}{Zhuoxiao Chen}, \bibinfo{person}{Sen Wang}, {and} \bibinfo{person}{Zi Huang}.} \bibinfo{year}{2024}\natexlab{}.
\newblock \showarticletitle{In search of lost online test-time adaptation: A survey}.
\newblock \bibinfo{journal}{\emph{International Journal of Computer Vision}} (\bibinfo{year}{2024}), \bibinfo{pages}{1--34}.
\newblock


\bibitem[Wu et~al\mbox{.}(2021)]%
        {wu2021multimodal}
\bibfield{author}{\bibinfo{person}{Yang Wu}, \bibinfo{person}{Pengwei Zhan}, \bibinfo{person}{Yunjian Zhang}, \bibinfo{person}{Liming Wang}, {and} \bibinfo{person}{Zhen Xu}.} \bibinfo{year}{2021}\natexlab{}.
\newblock \showarticletitle{Multimodal fusion with co-attention networks for fake news detection}. In \bibinfo{booktitle}{\emph{Findings of the association for computational linguistics: ACL-IJCNLP 2021}}. \bibinfo{pages}{2560--2569}.
\newblock


\bibitem[Xu et~al\mbox{.}(2023)]%
        {xu2023combating}
\bibfield{author}{\bibinfo{person}{Danni Xu}, \bibinfo{person}{Shaojing Fan}, {and} \bibinfo{person}{Mohan Kankanhalli}.} \bibinfo{year}{2023}\natexlab{}.
\newblock \showarticletitle{Combating misinformation in the era of generative AI models}. In \bibinfo{booktitle}{\emph{Proceedings of the 31st ACM International Conference on Multimedia}}. \bibinfo{pages}{9291--9298}.
\newblock


\bibitem[Xu et~al\mbox{.}(2022)]%
        {xu2022ava}
\bibfield{author}{\bibinfo{person}{Eric~Zhongcong Xu}, \bibinfo{person}{Zeyang Song}, \bibinfo{person}{Satoshi Tsutsui}, \bibinfo{person}{Chao Feng}, \bibinfo{person}{Mang Ye}, {and} \bibinfo{person}{Mike~Zheng Shou}.} \bibinfo{year}{2022}\natexlab{}.
\newblock \showarticletitle{Ava-avd: Audio-visual speaker diarization in the wild}. In \bibinfo{booktitle}{\emph{Proceedings of the 30th ACM International Conference on Multimedia}}. \bibinfo{pages}{3838--3847}.
\newblock


\bibitem[Yang et~al\mbox{.}(2024)]%
        {yang2024qwen2}
\bibfield{author}{\bibinfo{person}{An Yang}, \bibinfo{person}{Baosong Yang}, \bibinfo{person}{Beichen Zhang}, \bibinfo{person}{Binyuan Hui}, \bibinfo{person}{Bo Zheng}, \bibinfo{person}{Bowen Yu}, \bibinfo{person}{Chengyuan Li}, \bibinfo{person}{Dayiheng Liu}, \bibinfo{person}{Fei Huang}, \bibinfo{person}{Haoran Wei}, {et~al\mbox{.}}} \bibinfo{year}{2024}\natexlab{}.
\newblock \showarticletitle{Qwen2. 5 technical report}.
\newblock \bibinfo{journal}{\emph{arXiv preprint arXiv:2412.15115}} (\bibinfo{year}{2024}).
\newblock


\bibitem[Yang et~al\mbox{.}(2023a)]%
        {yang2023deep}
\bibfield{author}{\bibinfo{person}{Yang Yang}, \bibinfo{person}{Ran Bao}, \bibinfo{person}{Weili Guo}, \bibinfo{person}{De-Chuan Zhan}, \bibinfo{person}{Yilong Yin}, {and} \bibinfo{person}{Jian Yang}.} \bibinfo{year}{2023}\natexlab{a}.
\newblock \showarticletitle{Deep visual-linguistic fusion network considering cross-modal inconsistency for rumor detection}.
\newblock \bibinfo{journal}{\emph{Science China Information Sciences}} \bibinfo{volume}{66}, \bibinfo{number}{12} (\bibinfo{year}{2023}), \bibinfo{pages}{222102}.
\newblock


\bibitem[Yang et~al\mbox{.}(2023b)]%
        {yang2023dawn}
\bibfield{author}{\bibinfo{person}{Zhengyuan Yang}, \bibinfo{person}{Linjie Li}, \bibinfo{person}{Kevin Lin}, \bibinfo{person}{Jianfeng Wang}, \bibinfo{person}{Chung-Ching Lin}, \bibinfo{person}{Zicheng Liu}, {and} \bibinfo{person}{Lijuan Wang}.} \bibinfo{year}{2023}\natexlab{b}.
\newblock \showarticletitle{The dawn of lmms: Preliminary explorations with gpt-4v (ision)}.
\newblock \bibinfo{journal}{\emph{arXiv preprint arXiv:2309.17421}} \bibinfo{volume}{9}, \bibinfo{number}{1} (\bibinfo{year}{2023}), \bibinfo{pages}{1}.
\newblock


\bibitem[Zeng et~al\mbox{.}(2023)]%
        {zeng2023correcting}
\bibfield{author}{\bibinfo{person}{Zhi Zeng}, \bibinfo{person}{Mingmin Wu}, \bibinfo{person}{Guodong Li}, \bibinfo{person}{Xiang Li}, \bibinfo{person}{Zhongqiang Huang}, {and} \bibinfo{person}{Ying Sha}.} \bibinfo{year}{2023}\natexlab{}.
\newblock \showarticletitle{Correcting the bias: Mitigating multimodal inconsistency contrastive learning for multimodal fake news detection}. In \bibinfo{booktitle}{\emph{2023 IEEE International Conference on Multimedia and Expo (ICME)}}. IEEE, \bibinfo{pages}{2861--2866}.
\newblock


\bibitem[Zhang et~al\mbox{.}(2024a)]%
        {zhang2024mitigating}
\bibfield{author}{\bibinfo{person}{Litian Zhang}, \bibinfo{person}{Xiaoming Zhang}, \bibinfo{person}{Chaozhuo Li}, \bibinfo{person}{Ziyi Zhou}, \bibinfo{person}{Jiacheng Liu}, \bibinfo{person}{Feiran Huang}, {and} \bibinfo{person}{Xi Zhang}.} \bibinfo{year}{2024}\natexlab{a}.
\newblock \showarticletitle{Mitigating social hazards: Early detection of fake news via diffusion-guided propagation path generation}. In \bibinfo{booktitle}{\emph{Proceedings of the 32nd ACM International Conference on Multimedia}}. \bibinfo{pages}{2842--2851}.
\newblock


\bibitem[Zhang et~al\mbox{.}(2024b)]%
        {zhang2024reinforced}
\bibfield{author}{\bibinfo{person}{Litian Zhang}, \bibinfo{person}{Xiaoming Zhang}, \bibinfo{person}{Ziyi Zhou}, \bibinfo{person}{Feiran Huang}, {and} \bibinfo{person}{Chaozhuo Li}.} \bibinfo{year}{2024}\natexlab{b}.
\newblock \showarticletitle{Reinforced adaptive knowledge learning for multimodal fake news detection}. In \bibinfo{booktitle}{\emph{Proceedings of the AAAI conference on artificial intelligence}}, Vol.~\bibinfo{volume}{38}. \bibinfo{pages}{16777--16785}.
\newblock


\bibitem[Zhang et~al\mbox{.}(2024c)]%
        {zhang2024early}
\bibfield{author}{\bibinfo{person}{Litian Zhang}, \bibinfo{person}{Xiaoming Zhang}, \bibinfo{person}{Ziyi Zhou}, \bibinfo{person}{Xi Zhang}, \bibinfo{person}{Senzhang Wang}, \bibinfo{person}{Philip~S Yu}, {and} \bibinfo{person}{Chaozhuo Li}.} \bibinfo{year}{2024}\natexlab{c}.
\newblock \showarticletitle{Early detection of multimodal fake news via reinforced propagation path generation}.
\newblock \bibinfo{journal}{\emph{IEEE Transactions on Knowledge and Data Engineering}} (\bibinfo{year}{2024}).
\newblock


\bibitem[Zhang et~al\mbox{.}(2023)]%
        {zhang2023hierarchical}
\bibfield{author}{\bibinfo{person}{Qiang Zhang}, \bibinfo{person}{Jiawei Liu}, \bibinfo{person}{Fanrui Zhang}, \bibinfo{person}{Jingyi Xie}, {and} \bibinfo{person}{Zheng-Jun Zha}.} \bibinfo{year}{2023}\natexlab{}.
\newblock \showarticletitle{Hierarchical semantic enhancement network for multimodal fake news detection}. In \bibinfo{booktitle}{\emph{Proceedings of the 31st ACM International Conference on Multimedia}}. \bibinfo{pages}{3424--3433}.
\newblock


\bibitem[Zhao et~al\mbox{.}(2023)]%
        {zhao2023enhancing}
\bibfield{author}{\bibinfo{person}{Wanqing Zhao}, \bibinfo{person}{Yuta Nakashima}, \bibinfo{person}{Haiyuan Chen}, {and} \bibinfo{person}{Noboru Babaguchi}.} \bibinfo{year}{2023}\natexlab{}.
\newblock \showarticletitle{Enhancing Fake News Detection in Social Media via Label Propagation on Cross-modal Tweet Graph}. In \bibinfo{booktitle}{\emph{Proceedings of the 31st ACM International Conference on Multimedia}}. \bibinfo{pages}{2400--2408}.
\newblock


\bibitem[Zhu et~al\mbox{.}(2022)]%
        {zhu2022memory}
\bibfield{author}{\bibinfo{person}{Yongchun Zhu}, \bibinfo{person}{Qiang Sheng}, \bibinfo{person}{Juan Cao}, \bibinfo{person}{Qiong Nan}, \bibinfo{person}{Kai Shu}, \bibinfo{person}{Minghui Wu}, \bibinfo{person}{Jindong Wang}, {and} \bibinfo{person}{Fuzhen Zhuang}.} \bibinfo{year}{2022}\natexlab{}.
\newblock \showarticletitle{Memory-guided multi-view multi-domain fake news detection}.
\newblock \bibinfo{journal}{\emph{IEEE Transactions on Knowledge and Data Engineering}} \bibinfo{volume}{35}, \bibinfo{number}{7} (\bibinfo{year}{2022}), \bibinfo{pages}{7178--7191}.
\newblock


\bibitem[Zhu et~al\mbox{.}(2024)]%
        {zhu2024vision+}
\bibfield{author}{\bibinfo{person}{Ye Zhu}, \bibinfo{person}{Yu Wu}, \bibinfo{person}{Nicu Sebe}, {and} \bibinfo{person}{Yan Yan}.} \bibinfo{year}{2024}\natexlab{}.
\newblock \showarticletitle{Vision+ x: A survey on multimodal learning in the light of data}.
\newblock \bibinfo{journal}{\emph{IEEE Transactions on Pattern Analysis and Machine Intelligence}} (\bibinfo{year}{2024}).
\newblock


\bibitem[Zong et~al\mbox{.}(2024)]%
        {zong2024unveiling}
\bibfield{author}{\bibinfo{person}{Linlin Zong}, \bibinfo{person}{Jiahui Zhou}, \bibinfo{person}{Wenmin Lin}, \bibinfo{person}{Xinyue Liu}, \bibinfo{person}{Xianchao Zhang}, {and} \bibinfo{person}{Bo Xu}.} \bibinfo{year}{2024}\natexlab{}.
\newblock \showarticletitle{Unveiling opinion evolution via prompting and diffusion for short video fake news detection}. In \bibinfo{booktitle}{\emph{Findings of the Association for Computational Linguistics ACL 2024}}. \bibinfo{pages}{10817--10826}.
\newblock


\end{thebibliography}
\bibliographystyle{ACM-Reference-Format.bst}

\end{document}